\documentclass[conference]{IEEEtran}
\IEEEoverridecommandlockouts

\usepackage{amsmath,amssymb}
\usepackage{booktabs}
\usepackage{multirow}
\usepackage{tabularx}

\usepackage{cite}
\usepackage{amsmath,amssymb,amsfonts}
\usepackage{algorithmic}
\usepackage{graphicx}
\usepackage{textcomp}
\usepackage{xcolor}
\def\BibTeX{{\rm B\kern-.05em{\sc i\kern-.025em b}\kern-.08em
    T\kern-.1667em\lower.7ex\hbox{E}\kern-.125emX}}

\makeatletter
\newcommand{\linebreakand}{%
  \end{@IEEEauthorhalign}
  \hfill\mbox{}\par
  \mbox{}\hfill\begin{@IEEEauthorhalign}
}
\makeatother

\begin{document}

\title{Local Prototype Reconstruction for Text-Compatible Speech-to-LLM Bridge Pretraining \\
\thanks{Thanks to grant S004923N from Foundation Flanders (FWO).}
}




\author{
  \IEEEauthorblockN{
    Xinnian Zhao$^{\star}$,
    Chia-Hua Wu$^{\dagger}$,
    Pu Wang$^{\star}$,
    Hugo Van Hamme$^{\star}$
  }
  \IEEEauthorblockA{
    $^{\star}$Department of Electrical Engineering (ESAT), KU Leuven, Leuven, Belgium
  }
  \IEEEauthorblockA{
    $^{\dagger}$Institute of Information Science, Academia Sinica, Taiwan
  }
}

\maketitle

\begin{abstract}
Speech-to-LLM systems often connect a frozen speech encoder to a frozen large language model (LLM) through a small trainable bridge.
The bridge is usually treated as plumbing, but it in fact defines the geometry of the speech-to-LLM interface, and the pretraining objective decides whether that interface provides a reusable initialization for downstream tasks. We study a transferable bridge through two complementary properties: \emph{global alignment} with the text side, and \emph{local lexical manifold compatibility}, where bridge embeddings remain close to the frozen LLM's input-embedding neighbourhoods. 
We make this property measurable with a fixed, head-free, timestamp-free diagnostic that applies to any objective, and show that next-word prediction (NWP) and sentence-level contrastive pretraining do not fully capture token-level lexical compatibility.
We then introduce Local Prototype Reconstruction (LPR), a lightweight training-only regularizer that requires each aligned bridge token to be reconstructable from a small neighbourhood of frozen LLM token embeddings, with a hard single-prototype anchor as its limiting case. On multilingual ASR and speech translation, LPR improves transfer, with the largest gains on translation and low-resource adaptation. Crucially, our independent diagnostic correlates with downstream gains across objectives, 
suggesting that lexical manifold compatibility is predictive of reusability for speech-to-LLM bridges.
\end{abstract}


\section{Introduction}

Large language models (LLMs)~\cite{achiam2023gpt,bai2023qwen,anil2023palm} are increasingly extended with speech input through modality mapping: a frozen speech foundation encoder extracts acoustic representations, and a lightweight bridge projects them into the LLM embedding space for tasks such as automatic speech recognition (ASR) and speech translation (ST)~\cite{verdini2025connect,ma2025speech,xue2025ideal,hu2024gentranslate,gaido2024speech}. This bridge is often viewed as a small projection module whose role is simply to make speech readable to the frozen LLM. We take a more structural view: when both the speech encoder and the LLM are frozen, the bridge defines the geometry of the speech-to-LLM interface. This raises a central question: \emph{what pretraining objective yields a reusable speech interface, rather than a representation that is only tuned to one generation task?}

We study this question through two complementary properties. The first is \emph{global semantic alignment}: a pooled speech representation should match the corresponding text representation, so that utterance-level meaning is preserved. The second is \emph{local lexical manifold compatibility}: individual bridge embeddings should remain close to the local neighbourhoods of the frozen LLM input-embedding space. Our hypothesis is that global alignment is important but may be insufficient for a reusable bridge, because the LLM ultimately consumes a sequence of token-like embeddings rather than only a pooled utterance vector. We therefore make both properties measurable and study how different pretraining objectives affect them.

\begin{figure}[t]
\centering
\includegraphics[width=\columnwidth]{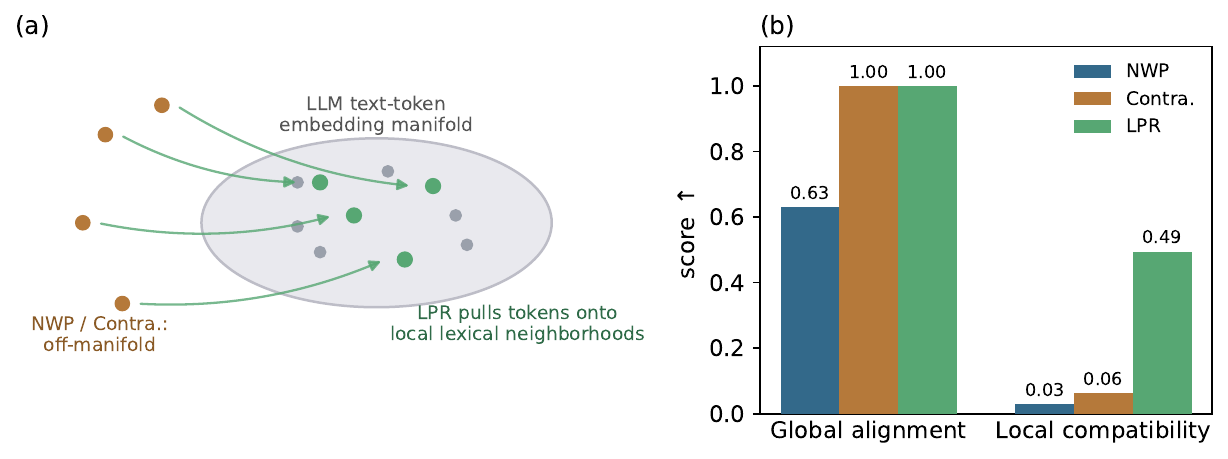}
\caption{Motivating gap between global alignment and local compatibility. (a) Schematic view: NWP and contrastive objectives can produce bridge embeddings that support generation or utterance-level alignment without necessarily matching the local neighbourhoods of the frozen LLM input-embedding manifold. LPR adds a token-level regularizer toward these neighbourhoods. (b) Fixed head-free probes on the frozen bridge. Contrastive and LPR both reach near-perfect utterance-level retrieval R@1 ($\approx1.0$), while token-level cosine to nearby LLM embeddings remains low for contrastive ($0.06$) and increases with LPR ($0.49$).}\vspace{-2mm}
\label{fig:1}
\vspace{-2mm}
\end{figure}

Viewed through this lens, common objectives emphasize different parts of the interface. Next-word prediction (NWP) inherits the autoregressive objective of text LLMs~\cite{wu2023decoder,ma2024embarrassingly,yu2024connecting,hu2024wavllm,zhang2025soundwave}; it trains the model to generate text from speech-conditioned inputs, but it does not directly specify how adapted speech tokens should relate to the LLM lexical embedding space. Sentence-level contrastive pretraining~\cite{zufle2025contrastive} more directly improves speech-text alignment, but it operates on pooled utterance representations and therefore leaves token-level compatibility under-constrained. Figure~\ref{fig:1} illustrates this separation: contrastive pretraining reaches 
near-perfect global retrieval score, yet remains much closer to NWP on a token-level local-compatibility probe. This motivates a pretraining objective that preserves global alignment while also shaping the local geometry of individual bridge embeddings.

Guided by this observation, we propose \textbf{Local Prototype Reconstruction (LPR)}: a continuous, local, training-only regularizer that requires each transcript-aligned bridge token to be reconstructable from a small neighbourhood of frozen LLM token embeddings. LPR keeps the bridge continuous rather than discretizing speech into text tokens, keeps sentence-level contrastive learning as the global alignment backbone, and adds a local constraint that encourages compatibility with the LLM lexical manifold. A hard single-prototype anchor toward the gold token appears as the limiting case of the same framework.

Our contribution is to identify local lexical manifold compatibility as a measurable property of the speech-to-LLM interface and to show that encouraging this property improves transfer. Specifically:
\begin{itemize}
\item \textbf{Measurement.} We introduce a fixed, head-free, timestamp-free diagnostic of interface compatibility that can be applied uniformly to NWP, contrastive, and LPR checkpoints.
\item \textbf{Method.} We propose LPR, a continuous local-manifold regularizer with the hard anchor as its single-prototype ($K{=}1$) limit, added on top of the contrastive backbone and used only during pretraining.
\item \textbf{Empirical evidence.} On multilingual ASR and ST, LPR improves transfer most under low-resource adaptation and on translation, and the independent compatibility diagnostic correlates with downstream gains across objectives.
\end{itemize}

\section{Related Work}

\textbf{SpeechLLM bridges and pretraining objectives.}
Recent audio- and speech-LLMs commonly connect a pretrained acoustic encoder to a pretrained LLM through a lightweight connector, such as a projection module, Q-Former, or adapter, and then train the connector with autoregressive or instruction-following objectives~\cite{tangsalmonn,add-qwenaudio2023,add-speechgpt2023,add-llasm2023,add-speechverse2024,add-audioflamingo2024,add-rubenstein2023audiopalm,add-blsp2023,wu2023decoder,yu2024connecting,ma2024embarrassingly,hu2024wavllm}. These systems demonstrate that frozen or partially frozen LLMs can consume speech-derived representations, but the connector is typically evaluated through downstream generation quality rather than through the geometry of the interface it creates. Recent work has begun to compare connector designs and pretraining objectives for SpeechLLMs~\cite{verdini2025connect,gaido2024speech,ma2025speech,xue2025ideal,zhang2025soundwave}. Closest to our setting, contrastive SpeechLLM pretraining aligns pooled speech and text representations and improves task transfer~\cite{zufle2025contrastive}. We use this sentence-level contrastive objective as the global-alignment backbone, but ask a complementary question: after an utterance is globally aligned, are the individual bridge embeddings locally compatible with the frozen LLM's lexical embedding space?

\textbf{Cross-modal alignment, speech tokenization, and lexical interfaces.}
Contrastive learning has been widely used to align paired modalities at the representation level, from image--text models such as CLIP~\cite{add-radford2021clip} to audio--text models such as CLAP~\cite{add-elizalde2023clap}. Such objectives are effective for global retrieval-style alignment, but they do not by themselves specify how each continuous speech token should relate to the token-level interface expected by an LLM. Other approaches make speech more language-model-like by mapping audio into discrete or near-discrete units, for example through HuBERT-style hidden units~\cite{add-hsu2021hubert}, AudioLM-style audio tokenization~\cite{add-borsos2023audiolm}, continuous/discrete hybrid audio representations~\cite{add-lauragpt2023}, or speech-to-speech/audio LLM systems that operate over learned audio tokens~\cite{add-rubenstein2023audiopalm,add-speechgpt2023}. Discretization provides an explicit token interface, but can discard useful continuous information and tie the bridge to a particular tokenizer. A related line aligns speech and text inside the LLM with optimal transport for translation~\cite{liu2025adaptive} and knowledge-distillation approaches to speech-text pretraining~\cite{add-blspkd2024}; these methods target hidden-state or output-distribution alignment, whereas LPR regularizes the bridge interface toward the frozen input-embedding manifold.

\textbf{Representation geometry and interface diagnostics.}
Our measurement framework is also related to work that probes the geometry of neural language representations. Prior studies show that contextual embedding spaces contain strong geometric structure and that similarity-based probes can reveal how representations cluster, specialize, or collapse~\cite{add-ethayarajh2019contextual}. We adapt this perspective to SpeechLLM bridges: rather than asking whether text embeddings are geometrically well behaved in isolation, we ask whether speech-derived bridge tokens occupy neighbourhoods that the frozen LLM already uses for lexical input. The resulting diagnostic is model-agnostic, head-free, and timestamp-free, so it can compare NWP, contrastive, and LPR checkpoints under the same probe. This distinguishes our contribution from prior bridge training methods: LPR is one intervention, while the diagnostic provides a reusable way to measure local lexical manifold compatibility and test whether that property predicts downstream transfer.

\section{Method}
\label{sec:method}

\begin{figure*}[!t]
    \centering
    \includegraphics[width=\linewidth]{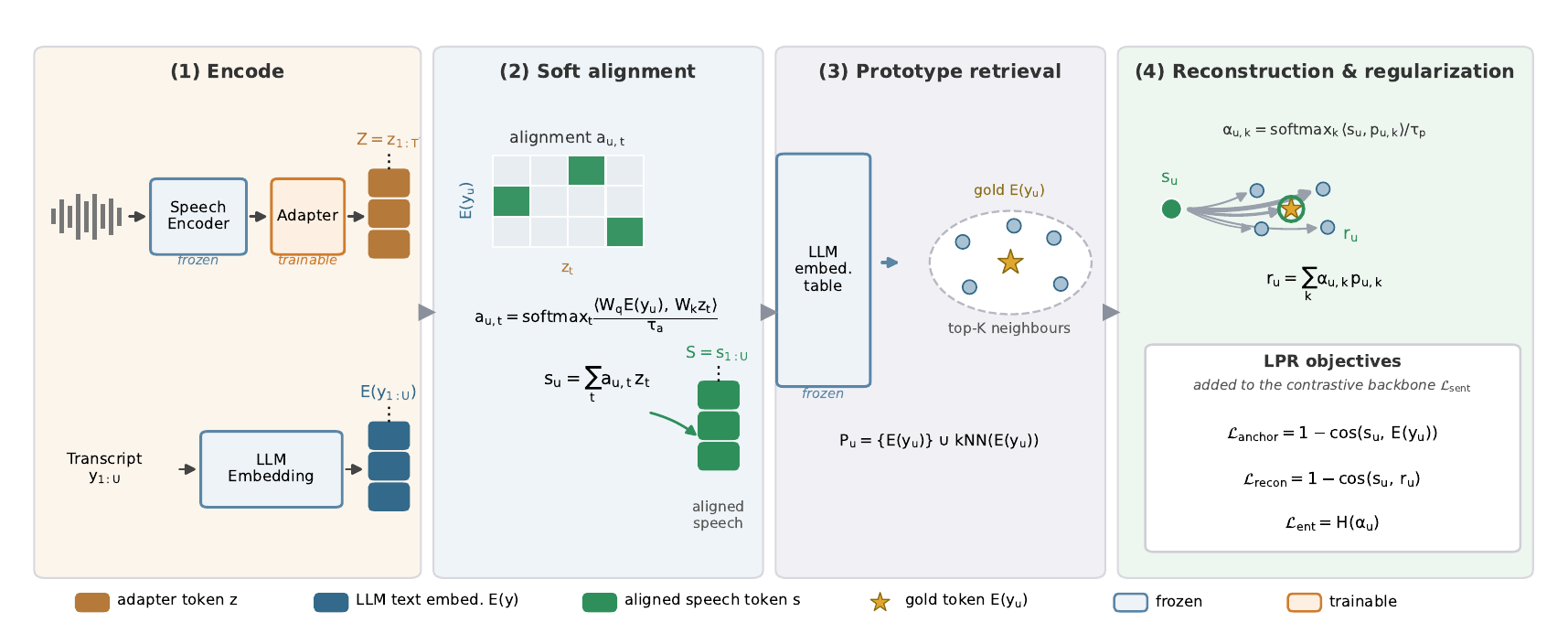}
    \caption{Overview of Local Prototype Reconstruction (LPR). During pretraining, transcript tokens are softly aligned to continuous bridge embeddings through a schematic soft-alignment module, local frozen-LLM embedding neighbourhoods are retrieved, and aligned speech representations are reconstructed from these prototypes. The resulting token-level losses refine the sentence-level contrastive bridge. At inference, only the encoder--bridge path is used; transcripts, prototype retrieval, and LPR heads are discarded.}
    \label{fig:2}
\end{figure*}

We study a Speech-to-LLM architecture in which a frozen speech encoder $\mathcal{S}$ is connected to a frozen LLM through a trainable bridge module $\mathcal{A}$. Let $E(\cdot)$ denote the frozen LLM input-embedding layer. Given speech $\mathbf{x}$ and transcript tokens $y_{1:U}$, the bridge maps encoder features into a sequence of continuous bridge embeddings
\begin{equation}
Z=z_{1:T'}=\mathcal{A}(\mathcal{S}(\mathbf{x}))\in\mathbb{R}^{T'\times d},
\end{equation}
while the transcript is embedded as
\begin{equation}
E(y_{1:U})=[E(y_1),\ldots,E(y_U)]\in\mathbb{R}^{U\times d}.
\end{equation}
The speech encoder and LLM are frozen throughout; only the bridge and the small LPR alignment projections are updated. Our objective combines utterance-level semantic alignment with token-level compatibility to the frozen LLM lexical embedding space.

\subsection{Sentence-Level Contrastive Backbone}

We use sentence-level contrastive pretraining to align speech and text globally. For each example $i$ in a mini-batch of size $N$, the LLM is run with either adapted speech tokens $Z_i$ or transcript embeddings $E(y_{i,1:U_i})$. Let $H^{(\ell)}_{a,i}$ and $H^{(\ell)}_{b,i}$ be the corresponding speech- and text-side hidden states at layer $\ell$, with $H^{(0)}_{a,i}=Z_i$ and $H^{(0)}_{b,i}=E(y_{i,1:U_i})$. After mean pooling and $\ell_2$ normalization, cosine similarities $s^{(\ell)}_{ij}$ are used in a symmetric InfoNCE loss:
\begingroup
\setlength{\abovedisplayskip}{0.5em}
\setlength{\belowdisplayskip}{0.5em}
\begin{equation}
\thickmuskip=0.2mu
\medmuskip=0.2mu
\mathcal{L}^{(\ell)}_{\mathrm{sent}}
= -\frac{1}{2N}\sum_{i=1}^{N}\left[
\log\frac{e^{s^{(\ell)}_{ii}/\tau_\ell}}{\sum_{j=1}^{N}e^{s^{(\ell)}_{ij}/\tau_\ell}}
+\log\frac{e^{s^{(\ell)}_{ii}/\tau_\ell}}{\sum_{j=1}^{N}e^{s^{(\ell)}_{ji}/\tau_\ell}}
\right],
\end{equation}
\endgroup
where $\tau_\ell$ is a learnable temperature. The backbone loss averages over selected LLM layers:
\begin{equation}
\mathcal{L}_{\mathrm{sent}}
=
\frac{1}{|\mathcal{L}_{\mathrm{layers}}|}
\sum_{\ell\in\mathcal{L}_{\mathrm{layers}}}
\mathcal{L}^{(\ell)}_{\mathrm{sent}} .
\end{equation}
This aligns pooled utterance representations, but does not directly constrain individual bridge outputs.

\subsection{Local Prototype Reconstruction}

LPR, shown in Figure~\ref{fig:2}, adds a token-level constraint on top of the contrastive backbone. The key idea is to keep the speech bridge continuous, but require each transcript-aligned bridge token to be explainable by a small local neighbourhood of frozen LLM token embeddings. This is softer than discretizing speech into text tokens, yet more structured than leaving bridge outputs as unconstrained continuous vectors.


Because the bridge sequence length $T'$ does not generally match the transcript length $U$, LPR first derives a transcript-conditioned soft alignment. We compute this alignment in a low-dimensional similarity space using bias-free projections $W_q,W_k\in\mathbb{R}^{d_a\times d}$, with $d_a=512$ in our experiments:
\begin{equation}
a_{u,t}
=
\mathrm{softmax}_{t}
\left(
\frac{\cos(W_q E(y_u), W_k z_t)}{\tau_a}
\right),
\qquad
s_u=\sum_{t=1}^{T'}a_{u,t}z_t .
\end{equation}
The projections are used only to compute alignment weights. The aligned representation $s_u$ remains a weighted sum of the original bridge outputs and is compared to $E(y_u)$ and its prototypes in the LLM input-embedding space. Thus the alignment head learns a low-rank similarity metric for soft assignment, not a separate space in which lexical compatibility is evaluated.

For each transcript position $u$, we retrieve a local prototype set
\begin{equation}
P_u=\{p_{u,1},\ldots,p_{u,K}\},\qquad p_{u,1}=E(y_u),
\end{equation}
where the remaining $K-1$ prototypes are nearest neighbours of $E(y_u)$ in the frozen LLM input-embedding table, ranked by cosine similarity between $\ell_2$-normalized embeddings. The candidate pool is built offline from the most frequent transcript tokens in the pretraining data and is kept fixed during training.
By restricting reconstruction to a neighbourhood of the gold token, LPR asks whether the speech representation lies near the relevant part of the lexical manifold rather than merely near the vocabulary as a whole.

We then reconstruct $s_u$ from this local neighbourhood:
\begin{equation}
\alpha_{u,k}
=
\mathrm{softmax}_{k}
\left(
\frac{\cos(s_u,p_{u,k})}{\tau_p}
\right),
\qquad
r_u=\sum_{k=1}^{K}\alpha_{u,k}p_{u,k}.
\end{equation}
The reconstruction $r_u$ is a local convex combination of frozen LLM token embeddings. It does not force $s_u$ to equal a discrete token; instead, it encourages $s_u$ to occupy a region that the LLM already uses as a lexical interface.

The token-level losses are
\begin{align}
\mathcal{L}_{\mathrm{anchor}}
&=
\frac{1}{U}\sum_{u=1}^{U}\left[1-\cos(s_u,E(y_u))\right],\\
\mathcal{L}_{\mathrm{recon}}
&=
\frac{1}{U}\sum_{u=1}^{U}\left[1-\cos(s_u,r_u)\right],
\end{align}
The anchor term provides direct lexical grounding to the gold token, while the reconstruction term tests whether this grounding extends to a local neighbourhood. We use cosine reconstruction to emphasize direction in the LLM embedding space rather than embedding norm. 
When $K=1$, $P_u=\{E(y_u)\}$, $\alpha_{u,1}=1$, and $r_u=E(y_u)$, so $\mathcal{L}_{\mathrm{recon}}$ reduces to $\mathcal{L}_{\mathrm{anchor}}$. Anchor and reconstruction are therefore hard and soft points on the same lexical-grounding axis.

\subsection{Training Objective and Inference}

The full pretraining objective is
\begin{equation}
\mathcal{L}
=
\mathcal{L}_{\mathrm{sent}}
+
\lambda_a \mathcal{L}_{\mathrm{anchor}}
+
\lambda_r \mathcal{L}_{\mathrm{recon}}
\end{equation}
Setting $\lambda_a=\lambda_r=0$ 
recovers the contrastive baseline with the same architecture and training pipeline. Transcript-conditioned alignment, prototype retrieval, and reconstruction are used only to compute pretraining regularizers. During downstream fine-tuning and inference, the model uses only the frozen speech encoder, trained bridge, and frozen LLM.

\section{Measuring Interface Compatibility}
\label{sec:probe}

To compare objectives under a common lens, we use a fixed, head-free probe that is independent of any trained LPR alignment head. Given a frozen bridge and transcript, the probe forms transcript-aligned speech representations with a parameter-free soft alignment:
\begin{equation}
\tilde{a}_{u,t}
=
\mathrm{softmax}_{t}
\left(
\frac{\cos(E(y_u),z_t)}{\tau_{\mathrm{probe}}}
\right),
\qquad
\tilde{s}_u
=
\sum_t \tilde{a}_{u,t}z_t .
\end{equation}
It then retrieves the same local prototype set $P_u$ and reconstructs $\tilde{s}_u$ as in Section~\ref{sec:method}. The probe requires transcripts but no word timestamps, and is applied identically to NWP, contrastive, and LPR checkpoints.

We report five diagnostics:
\begin{itemize}
    \item \textbf{Utterance Retrieval R@1} measures global alignment by asking whether pooled speech and text representations retrieve their paired utterance among in-batch candidates.
    \item 
    \textbf{Manifold Reconstruction Error (MRE)} is $\|\hat{s}_u-\hat{r}_u\|_2^2$, the squared distance between a unit-normalized aligned speech token $\hat{s}_u$ and prototype reconstructions $\hat{r}_u$. Lower MRE means better reconstruction from the local LLM lexical neighbourhood.
    \item \textbf{Gold-token Rank} reports the R@1 and Mean Reciprocal Rank (MRR) of the gold embedding $E(y_u)$ among prototypes ranked by similarity to $\tilde{s}_u$, testing whether the aligned speech token identifies its corresponding lexical anchor.
    \item \textbf{Prototype Entropy $H(\alpha)$} measures whether reconstruction uses a compact local support or a diffuse mixture of prototypes.
    \item \textbf{Alignment Entropy $H(\tilde{a})$} measures whether each transcript token is associated with a localized or broad span of bridge outputs.
\end{itemize}

These diagnostics are used for analysis rather than baseline training. Since some quantities are related to LPR's objective, we interpret them together with downstream transfer rather than in isolation. The central question is whether a fixed diagnostic, applied uniformly across objectives, predicts ASR and ST gains.

\section{Experimental Setup}

We evaluate on CoVoST2~\cite{wang2020covost}, a multilingual speech translation benchmark derived from CommonVoice~\cite{ardila2020common}. We use five source languages with English targets, De/Fr/Es/It/Pt$\rightarrow$En, totaling approximately 431 hours for training, 83 hours for development, and 88 hours for test. The same pretrained bridge is fine-tuned for ASR, where the target is the source-language transcript, and ST, where the target is the English translation. We also evaluate a 10\% low-resource fine-tuning subset, corresponding to approximately 43 hours of speech.

Our model uses the encoder of \texttt{Whisper-large-v3}~\cite{radford2023robust} as the frozen speech encoder and \texttt{Llama-3-8B-Instruct}~\cite{grattafiori2024llama} as the frozen LLM. The trainable bridge is a window-level Q-Former~\cite{tangsalmonn} with 2 Transformer blocks, hidden size 4096, 8 attention heads, and window size 10, totaling approximately 177M trainable bridge parameters. We compare no pretraining, NWP, sentence-level contrastive pretraining, and LPR variants built on the contrastive backbone. 

For each objective, we first pre-train a single multilingual bridge on the five source languages. Downstream ASR and ST models are then fine-tuned separately from this same pretrained bridge initialization. Fine-tuning is also multilingual: all five languages are trained jointly, and the task and language are specified by prompts. The ``None'' pretraining baseline skips bridge pretraining. It is randomly initialized and trained only during downstream supervised fine-tuning. Since Llama is not pretrained to consume Whisper encoder states directly, the bridge is always required to map speech features into the LLM embedding dimension.

All objectives use the same data splits, bridge architecture, precision setting, and compute budget. Pretraining runs for up to 6 epochs on a single NVIDIA H100 80GB GPU with batch size 32. Downstream fine-tuning uses autoregressive training, capped at 80{,}000 steps with early stopping (patience 3). We use AdamW with peak learning rate $3\times10^{-5}$, warmup ratio 0.03, and beam search with beam size 3 for evaluation.

Unless otherwise stated, 
LPR uses bias-free query/key alignment projections from $d=4096$ to $d_a=512$.
$K=16$ local prototypes from an offline frozen bank, and alignment/prototype temperatures $\tau_a=\tau_p=0.1$. 
The fixed diagnostic probe uses $\tau_{\mathrm{probe}}=0.1$. 
LPR (Anch.) uses $\lambda_a=0.5,\lambda_r=0$, LPR (Recon.) uses $\lambda_a=0,\lambda_r=0.5$, and LPR (full) uses $\lambda_a=\lambda_r=0.5$. The contrastive baseline is the identical pipeline with all LPR weights set to zero.

\newcolumntype{Y}{>{\centering\arraybackslash}X}
\begin{table*}[t]
\centering
\caption{Downstream ASR and ST across pretraining objectives (per-language and average). Each row uses one multilingual bridge initialization. LPR performs best in all columns; gains are largest on ST with low-resource adaptation. Best per block in bold.}
\renewcommand{\arraystretch}{1.05}
\setlength{\aboverulesep}{1pt}
\setlength{\belowrulesep}{1pt}
\begin{tabularx}{\linewidth}{c|l|*{6}{Y}|*{6}{Y}}
\noalign{\hrule height 1pt}
\multirow{3}{*}{Data} & \multirow{3}{*}{Pretrain Obj.} & \multicolumn{6}{c|}{ASR} & \multicolumn{6}{c}{ST} \\ \cline{3-14}
 &  & \multicolumn{6}{c|}{WER (\%) $\downarrow$} & \multicolumn{6}{c}{BLEU $\uparrow$} \\ \cline{3-14}
 &  & De & Fr & Es & It & Pt & Avg. & De$\to$En & Fr$\to$En & Es$\to$En & It$\to$En & Pt$\to$En & Avg. \\ \hline
\multirow{6}{*}{$\sim$430h} & None & 7.97 & 10.60 & 5.61 & 6.94 & 5.34 & 7.75 & 35.24 & 39.02 & 41.08 & 37.33 & 49.68 & 39.09 \\
 & NWP & 7.97 & 10.58 & 5.54 & 6.73 & 5.11 & 7.67 & 35.17 & 38.97 & 41.02 & 37.18 & 49.25 & 38.99 \\
 & Contra. & 7.94 & 10.52 & 5.43 & 6.41 & 4.98 & 7.56 & 37.67 & 39.93 & 41.82 & 38.28 & 50.90 & 40.36 \\
& LPR (Anch.) & 7.73 & 10.51 & 5.42 & 6.11 & 4.74 & 7.43 & 38.29 & 40.53 & 42.32 & 38.95 & 51.28 & 40.95 \\
& LPR (Recon.) & 7.69 & 10.41 & 5.36 & 6.07 & 4.55 & 7.36 & 38.29 & 40.60 & 42.35 & 39.07 & 51.65 & 41.02 \\
 & LPR (full) & \textbf{7.52} & \textbf{10.26} & \textbf{5.16} & \textbf{6.02} & \textbf{4.46} & \textbf{7.22} & \textbf{38.32} & \textbf{40.84} & \textbf{42.37} & \textbf{39.15} & \textbf{51.68} & \textbf{41.11} \\
 \hline
\multirow{6}{*}{$\sim$43h} & None & 10.92 & 13.55 & 7.61 & 9.28 & 7.18 & 10.28 & 31.60 & 35.08 & 37.18 & 33.49 & 45.55 & 35.24 \\
 & NWP & 10.91 & 13.56 & 7.59 & 9.27 & 7.03 & 10.26 & 28.95 & 32.21 & 35.27 & 31.05 & 42.04 & 32.68 \\
 & Contra. & 10.84 & 13.51 & 7.44 & 9.46 & 6.84 & 10.21 & 33.32 & 35.91 & 38.60 & 34.87 & 46.74 & 36.55 \\
 & LPR (Anch.) & 9.66 & 12.59 & 7.01 & 8.30 & 5.91 & 9.31 & 36.47 & 38.79 & 40.80 & 37.53 & 49.88 & 39.32 \\
 & LPR (Recon.) & 9.60 & 12.49 & 6.93 & 8.23 & 5.82 & 9.23 & 36.48 & 38.81 & 40.94 & 37.57 & 49.95 & 39.37 \\
 & LPR (full) & \textbf{9.50} & \textbf{12.40} & \textbf{6.72} & \textbf{8.07} & \textbf{5.52} & \textbf{9.08} & \textbf{36.60} & \textbf{38.94} & \textbf{40.96} & \textbf{37.76} & \textbf{50.22} & \textbf{39.49} \\
\noalign{\hrule height 1pt}
\end{tabularx}
\label{tab:1}
\vspace{-3mm}
\end{table*}

\section{Results}

\begin{figure}[!thbp]
\centering
\includegraphics[width=\columnwidth]{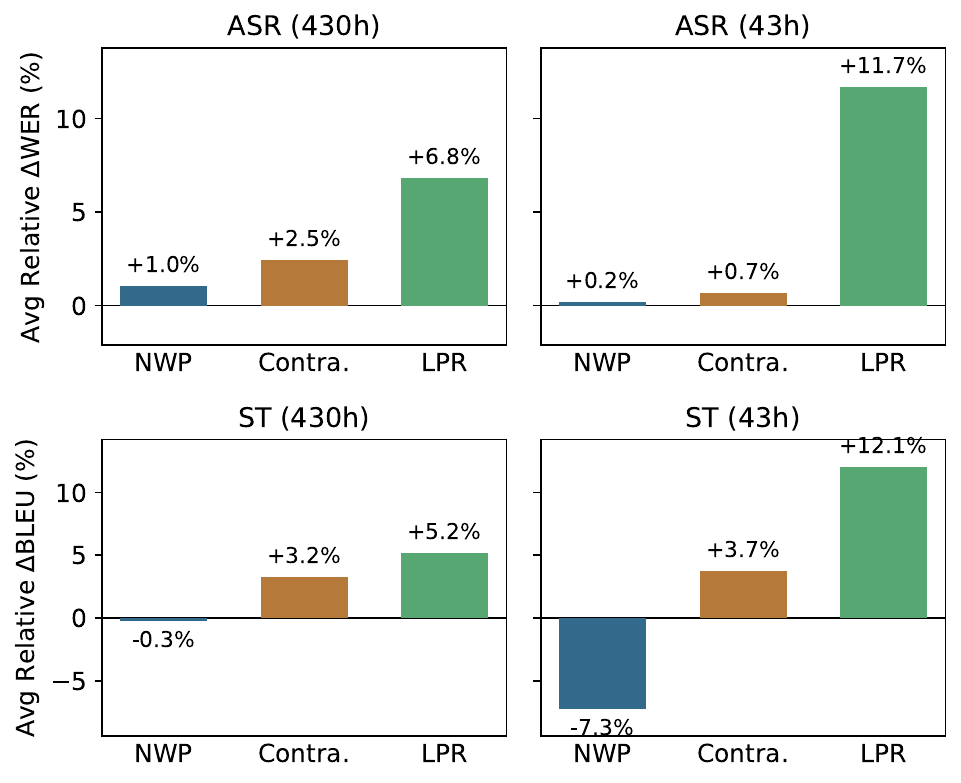}
\caption{Average relative downstream improvement over no pretraining across five source languages. ASR improvement is measured as relative WER reduction; ST improvement is measured as relative BLEU gain. LPR gives the largest average improvement in both tasks, with a larger margin in the low-resource setting.}\vspace{-2mm}
\label{fig:3}
\vspace{-2mm}
\end{figure}

Table~\ref{tab:1} reports downstream ASR (WER$\downarrow$) and ST (BLEU$\uparrow$) under full-resource ($\sim$430h) and low-resource ($\sim$43h) adaptation. Figure~\ref{fig:3} summarizes the average relative gain over the no-pretraining baseline. 
For each objective, ASR and ST fine-tuning start from the same pretrained multilingual bridge initialization. The downstream models are fine-tuned separately, so the comparison measures how well each pretraining objective provides a reusable bridge initialization across tasks.

\textbf{ASR.}
The ASR gains are moderate but consistent. Under full-resource adaptation, average WER improves from 7.75\% without pretraining to 7.67\% with NWP, 7.56\% with contrastive pretraining, and 7.22\% with LPR (full), corresponding to relative WER reductions of 1.0\%, 2.5\%, and 6.8\%. Under low-resource adaptation, the gap widens: LPR reduces average WER from 10.28\% to 9.08\%, while NWP and contrastive give smaller reductions. The trend is also consistent across the five source languages, with LPR (full) giving the best WER in each language and data regime. This suggests that even for transcription, where the target is close to the source-language surface form, the structure learned during bridge pretraining becomes more useful when downstream supervision is limited.

\textbf{ST.}
The differences are larger for speech translation. At full resource, NWP is close to the no-pretraining baseline, contrastive pretraining improves average BLEU from 39.09 to 40.36, and LPR further improves it to 41.11. In the low-resource setting, the separation is more pronounced: NWP falls below the no-pretraining baseline, contrastive reaches 36.55 BLEU, and LPR reaches 39.49 BLEU. The larger gain on ST is important because translation requires the bridge to support a target distribution different from the source transcript, rather than only preserving source-language surface form. This pattern is consistent with the view that NWP mainly optimizes generation from speech-conditioned inputs, whereas translation benefits more from a bridge that transfers across target distributions.

\textbf{Effect of limited adaptation data.}
The strongest gains appear where reuse of the pretrained interface should matter most. LPR improves both ASR and ST from the same pretrained checkpoint, and its margin over contrastive pretraining increases under the 10\% low-resource setting, especially for ST. This is the setting in which downstream fine-tuning has less capacity to repair a weak interface, so differences introduced during pretraining are more exposed. These results do not by themselves prove that local lexical compatibility is the only factor, but they motivate the analysis below: a better reusable bridge should improve downstream transfer most when fine-tuning data is scarce.

\section{Analysis: Why It Transfers}
\label{sec:analysis}

\begin{table}[t]
\centering
\caption{Interface-compatibility diagnostics on the multilingual development set, computed with the fixed head-free probe. Utterance retrieval measures global alignment; the remaining metrics measure token-level lexical manifold compatibility.}\vspace{-2mm}
\label{tab:diag}
\setlength{\tabcolsep}{3.0pt}
\begin{tabular}{lcccccc}
\toprule
Pretrain Obj. & Utt.\,R@1$\uparrow$ & Gold\,R@1$\uparrow$ & MRR$\uparrow$ & MRE$\downarrow$ & $H(\alpha)\!\downarrow$ & $H(a)\!\downarrow$ \\
\midrule
NWP     & 0.631 & 0.081 & 0.260 & 1.887 & 2.763 & 3.338 \\
Contra. & \textbf{0.999} & 0.376 & 0.573 & 1.835 & 2.754 & 3.323 \\
\midrule
\;LPR (Anch.)  & \textbf{0.999} & \textbf{0.553} & \textbf{0.708} & 0.718 & \textbf{1.841} & 1.831 \\
\;LPR (Recon.) & \textbf{0.999} & 0.209 & 0.430 & 0.760 & 2.392 & 2.219 \\
\;LPR (full)   & \textbf{0.999} & 0.443 & 0.611 & \textbf{0.645} & 1.912 & \textbf{1.822} \\
\bottomrule
\end{tabular}
\vspace{-2mm}
\end{table}

\textbf{Global alignment and local compatibility separate.}
Table~\ref{tab:diag} applies the fixed probe of Section~\ref{sec:probe} to the multilingual development set. 
Utterance-level retrieval shows that contrastive pretraining already solves the global alignment probe: contrastive and all LPR variants reach nearly perfect Utt.\,R@1, while NWP remains lower. This is not inconsistent with NWP's downstream performance, since this probe measures pooled speech-text retrieval rather than conditional generation quality.
The token-level metrics show a different pattern. Contrastive improves gold-token retrieval relative to NWP, but its MRE and entropy values remain close to NWP, indicating that strong utterance retrieval does not necessarily place bridge outputs near local LLM lexical neighbourhoods. LPR variants substantially reduce MRE and sharpen alignment entropy, suggesting that the added token-level regularization changes a property not captured by the sentence-level contrastive objective. The table also clarifies why global retrieval alone is not a sufficient diagnostic for bridge quality: once Utt.\,R@1 saturates, it cannot distinguish contrastive from LPR, even though their downstream ST and low-resource behavior differ.

\begin{figure}[t]
\centering
\includegraphics[width=\columnwidth]{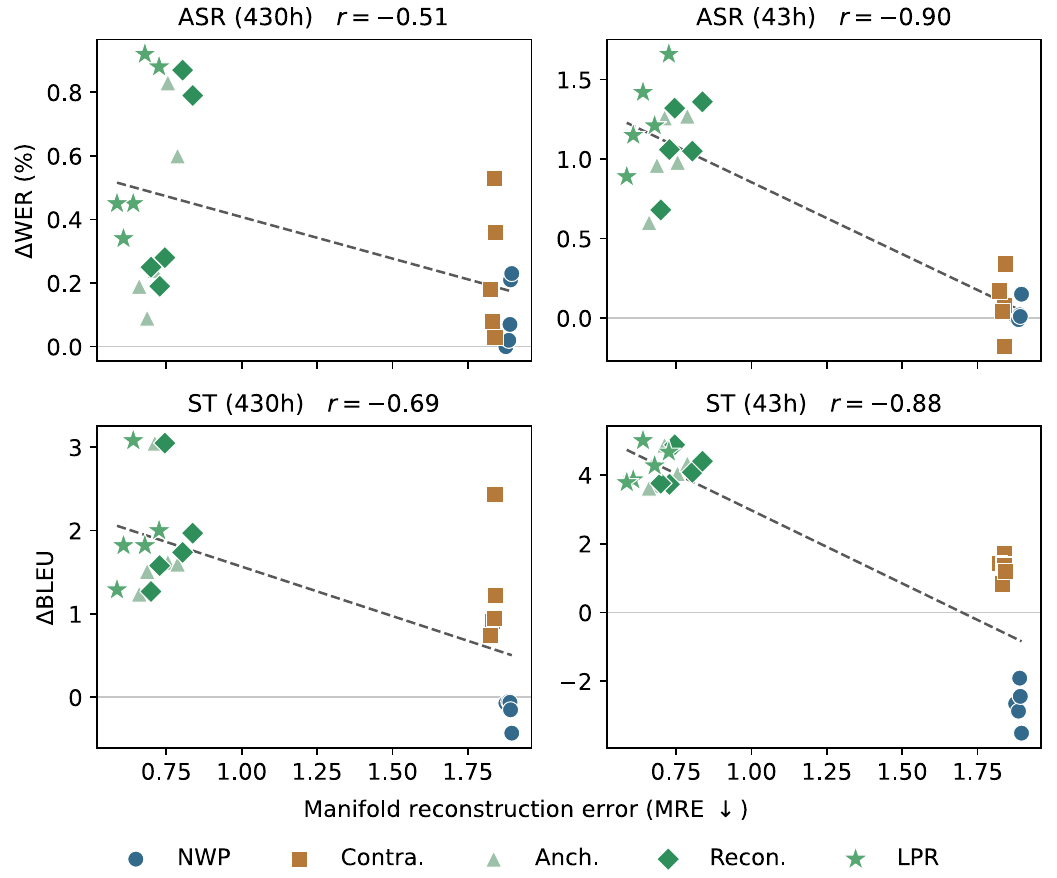}
\caption{Relationship between lexical manifold compatibility and downstream transfer. Each point is one objective-language pair. The x-axis shows MRE from the fixed head-free probe (lower is closer to the local LLM lexical manifold); the y-axis shows gain over no pretraining (ASR: WER reduction; ST: BLEU gain). Dashed lines are OLS fits and $r$ denotes Pearson correlation ($n=25$). The association is strongest in the low-resource setting.}\vspace{-2mm}
\label{fig:4}
\end{figure}

\textbf{Interface diagnostics correlate with downstream transfer.}
Figure~\ref{fig:4} plots, for each objective-language pair, the fixed probe's MRE against downstream gain over no pretraining. Lower MRE is associated with larger gains, with moderate correlations under full-resource adaptation (ASR $r{=}{-}0.51$, ST $r{=}{-}0.69$) and stronger correlations under low-resource adaptation (ASR $r{=}{-}0.90$, ST $r{=}{-}0.88$). The low-resource setting is especially informative because downstream fine-tuning has less capacity to repair a weak interface. Importantly, the trend includes NWP and contrastive checkpoints, which do not optimize MRE. Thus the diagnostic is not only showing that LPR moves a related training signal; it also ranks independently trained objectives in a way that tracks downstream transfer. The stronger correlation for ST and low-resource adaptation matches the pattern in Table~\ref{tab:1}: local lexical compatibility matters most when the bridge must generalize beyond transcription or when limited data constrains task-specific reshaping.

\section{Discussion}

The results clarify why the proposed diagnostic complements downstream
metrics. Downstream ASR and ST show whether a pretrained bridge
transfers, but not which part of the interface changed. The fixed
probe separates two properties that are often conflated:
utterance-level semantic alignment and token-level compatibility with
the frozen LLM lexical space. Contrastive pretraining captures the
first property well, while LPR mainly improves the second. This
distinction helps explain why objectives can look similar under global
retrieval yet behave differently under low-resource adaptation and
speech translation.

The correlation analysis further suggests that local lexical manifold
compatibility is not merely a diagnostic artifact. Although some probe
quantities relate to the LPR training signal, the same fixed probe is
applied to all objectives, including NWP and contrastive checkpoints
that never optimize LPR losses. MRE tracks downstream gain most
strongly in the low-resource setting, where the pretrained bridge has
less opportunity to be reshaped by the task and its compatibility with
the frozen LLM becomes more consequential.

The anchor and reconstruction variants also show that local
compatibility is not a single scalar property. A hard anchor gives the
strongest gold-token retrieval because it directly pulls aligned
speech tokens toward the transcript embedding. Soft prototype
reconstruction is less tied to exact token retrieval and instead
encourages compatibility with a local neighbourhood of the lexical
manifold. Their combination gives the best downstream results in our
experiments, suggesting that direct lexical grounding and
neighbourhood-level reconstruction provide complementary constraints.

Finally, LPR is a training-time regularizer. It uses transcripts during
supervised pretraining to form soft alignments and retrieve local
prototypes, but the deployed model uses only the frozen speech encoder,
trained bridge, and frozen LLM. This keeps inference unchanged relative
to the contrastive baseline.

\section{Conclusion}

We studied the speech-to-LLM bridge as an interface whose geometry
affects transfer. Sentence-level contrastive pretraining provides
strong global alignment, but our fixed diagnostic shows that this does
not fully determine whether bridge outputs are locally compatible with
the frozen LLM lexical embedding space. LPR adds this token-level
constraint through a continuous, training-only prototype
reconstruction regularizer. Across multilingual ASR and speech
translation, LPR improves transfer, especially under low-resource
adaptation, and the independent compatibility diagnostic correlates
with these gains. These findings suggest that local lexical manifold
compatibility is a useful property to measure and encourage when
pretraining reusable SpeechLLM bridges.

\newpage


\bibliographystyle{IEEEtran}
\bibliography{mybib}

\end{document}